%% file: iclr2023_conference_tinypaper_revision.tex
\documentclass{article} 
\usepackage{iclr2023_conference_tinypaper,times}

\input{math_commands.tex}

\usepackage{hyperref}
\usepackage{url}
\usepackage{graphicx}
\usepackage{subcaption}
\usepackage{amssymb}
\usepackage{tikz,pgflibraryshapes}
\usetikzlibrary{shadows}
\usetikzlibrary{arrows}

\usepackage[noabbrev]{cleveref}
\newtheorem{theorem}{Theorem}[section]

\DeclareMathOperator*{\concat}{%
    \mathchoice%
        {\big\Vert}%
        {\big\Vert}%
        {\Vert}%
        {\Vert}%
}
\definecolor{tabblue}{HTML}{1f77b4}
\definecolor{tabgreen}{HTML}{2ca02c}
\graphicspath{{figs}}
\title{Nonlinear model reduction for operator learning}

\author{Hamidreza Eivazi, Stefan Wittek \& Andreas Rausch\\
ISSE, Technical University of Clausthal, 38678 Clausthal-Zellerfeld, DE\\
\texttt{\{he76,stefan.wittek,andreas.rausch\}@tu-clausthal.de}
}

\iclrfinalcopy 
\begin{document}

\maketitle

\begin{abstract}
   Operator learning provides methods to approximate mappings between infinite-dimensional function spaces. Deep operator networks (DeepONets) are a notable architecture in this field. Recently, an extension of DeepONet based on model reduction and neural networks, proper orthogonal decomposition (POD)-DeepONet, has been able to outperform other architectures in terms of accuracy for several benchmark tests. We extend this idea towards nonlinear model order reduction by proposing an efficient framework that combines neural networks with kernel principal component analysis (KPCA) for operator learning. Our results demonstrate the superior performance of KPCA-DeepONet over POD-DeepONet.
\end{abstract}

\section{Introduction}
\paragraph{Operator learning.}
Partial Differential Equations (PDEs) are a fundamental mathematical tool for describing and analyzing physical phenomena that evolve in time and space \citep{brunton2023machine}. Solving the so-called parametric PDEs requires repeated operation of an expensive forward model (e.g. finite element methods \citep{FEM}) for every instance of the PDE, which demands a formidable cost. Recently, a new branch of ML research (the so-called operator learning) has made substantial advances for solving parametric PDEs by providing methods for learning operators, i.e. maps between infinite-dimensional spaces \citep{kovachki2023neural}. Operator networks are, by construction, resolution-independent; the model can provide solutions for any arbitrary input coordinate. DeepONet \citep{lu2019deeponet,lu2021learning} and its POD-based extension (POD-DeepONet) \citep{lu2022poddeeponet}, Fourier neural operator (FNO) \citep{li2020fourier}, and PCA-based neural networks (PCANN) \citep{bhattacharya2020model} are among successful operator learning approaches. 

\paragraph{Our contributions.} POD-DeepONet proposed by \cite{lu2022poddeeponet} follows the idea of PCANN \citep{bhattacharya2020model} for employing POD to represent functions. In POD-DeepONet, the trunk network of the DeepONet is replaced by a set of pre-computed POD bases obtained from the training data. However, POD is a linear decomposition technique and its ability to represent functions may be limited, especially for complex high-dimensional functions. In this contribution, we introduce kernel-PCA DeepONet (KPCA-DeepONet) as an efficient framework to combine nonlinear model reduction with operator learning. 
\begin{itemize}
   \item The KPCA-DeepONet is the first work that benefits from kernel methods and nonlinear model reduction techniques for learning operators.
   \item KPCA-DeepONet provides a non-linear reconstruction using kernel ridge regression.
   \item Our method provides less than 1\% error, the lowest error reported in the literature, for the benchmark test case of the Navier--Stokes equation.
\end{itemize}

\section{Methodology}

Let us consider $\gU$ and $\gV$ as two separable Banach spaces and assume that $\gG: \gU \mapsto \gV$ is an arbitrary (possibly nonlinear) operator. We consider a setting in which we only have access to partially observed input/output data $\{u_i, v_i\}_{i=1}^N$ as $N$ elements of $\,\gU \times \gV$ such that 
\begin{equation}
   \gG(u_i) = v_i, \qquad \text{ for }i= 1, \cdots, N.
\end{equation}
The input function $u$ is defined on the domain $D \subset \mathbb{R}^q$ and the output function $v$ is defined on the domain $D^{\prime} \subset \mathbb{R}^{q^{\prime}}$. Moreover, we consider $\gP$ and $\gQ$ as two linear and bounded evaluation operators such that
\begin{equation}
   \gP: u \mapsto (u(\vx_1), u(\vx_2), \cdots, u(\vx_n))^T \quad \text{ and } \quad \gQ: v \mapsto (v(\vy_1), v(\vy_2), \cdots, v(\vy_m))^T,
\end{equation}
where $u(\vx_i) \in \sR$, $v(\vy_i) \in \sR$, and $\{\vx_i\}_{i=1}^n$ and $\{\vy_i\}_{i=1}^m$ indicate two sets of collocation points in the domains $D$ and $D^{\prime}$, respectively. Considering $U_i = \gP(u_i)$ and $V_i = \gQ(v_i)$, our goal is to learn an approximation of $\gG$ from the training dataset $\{U_i, V_i\}_{i=1}^N$. We refer to \cref{app:deeponets} and \cite{lu2022poddeeponet} for details on DeepONet and POD-DeepONet.

\paragraph{KPCA-DeepONet.} A diagram of our method is depicted in \cref{app:diagram} \cref{fig:diagram}. Let $k_v$, $k_z$ be kernel functions. The KPCA bases of the output function $v$ are computed by performing eigendecomposition on $\mK_v = k_v(V_i, V_j)$ obtained from the training data. We denote the coefficients of the first $p$ KPCA basis of $v_i$ as $\vz_i \in \mathbb{R}^{p}$. The reconstruction of $v_i$ from $\vz_i$ is obtained through a kernel ridge ($\ell_2$-regularized) regression $h: \vz \mapsto v$ following the representer theorem (\cref{app:theorem1}) for achieving an optimal solution \citep{learningwithKernels}. The branch network learns the mapping from the input function $U_i$ to the coefficients of the KPCA basis $\vz_i$ from data. Thus, the output of KPCA-DeepONet can be written as
\begin{equation}
   \gG(u)(\vy)\approx \sum_{i = 1}^N \alpha_i(\vy)\, k_z(\!\concat_{k=1}^{p}\!\!b_k(U), \vz_i^t) + \phi_0(\vy),
\end{equation}
where $N$ denotes the number of training samples, $\{b_1, b_2, \cdots, b_p\}$ are the $p$ outputs of the branch net, $\vz_i^t$ is the projection of the $i$-th output function $V_i$ of the training data on the $p$ KPCA bases, and $\phi_0$ is the mean function. $\alpha_i$ are the weights of the kernel ridge regression. $\Vert$ indicates concatenation. Similar to ideas in POD-DeepONet and PCANN, we interpolate the coefficients of the kernel ridge regression to obtain $\alpha_i(\vy)$ and satisfy the discretization-invariance of the output function. The KPCA bases are only required for training.

\section{Numerical experiments}

\begin{figure}[t]
   \centering
   \includegraphics[height=0.175\linewidth]{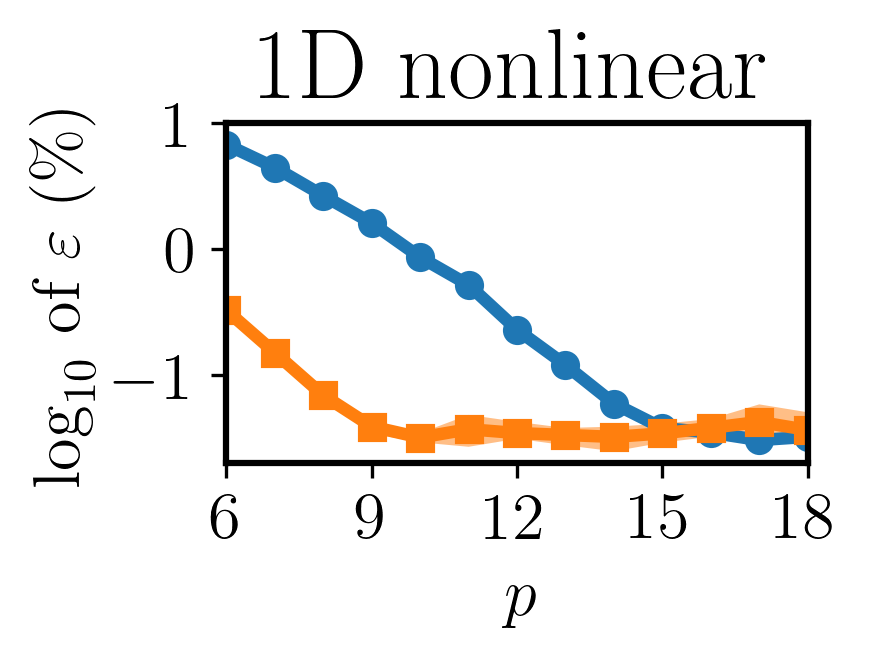}$\qquad$
   \includegraphics[height=0.175\linewidth]{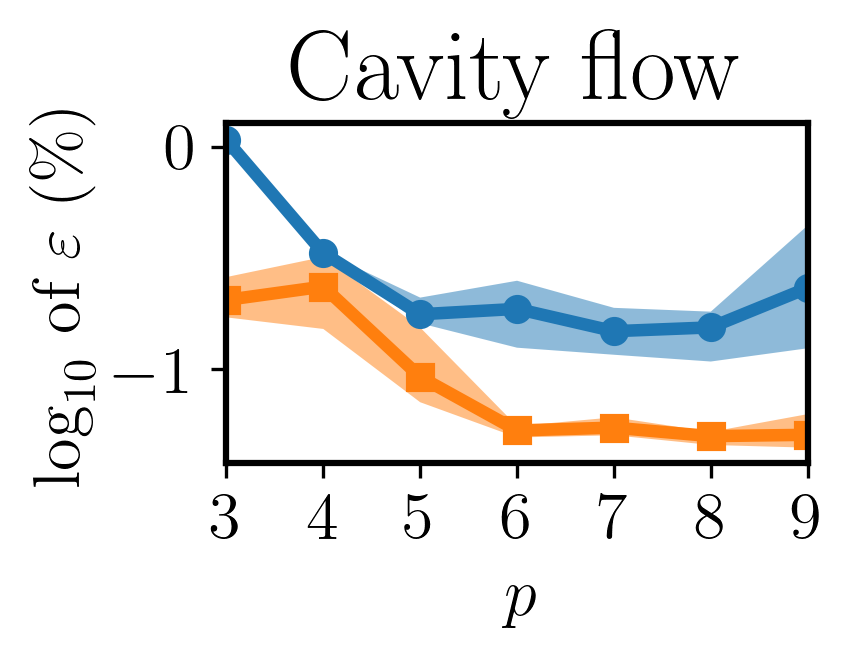}$\qquad$
   \includegraphics[height=0.175\linewidth]{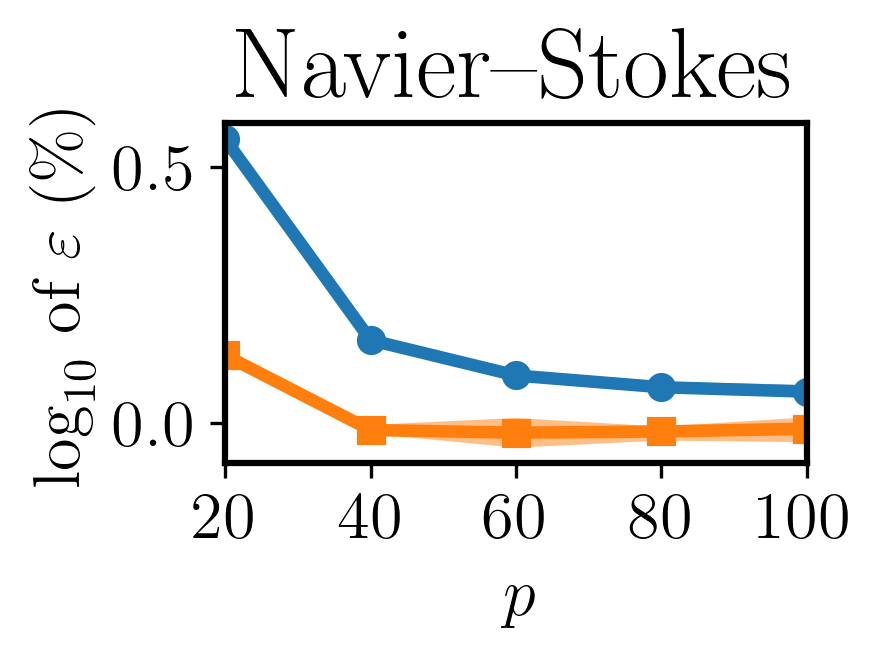}
   \caption{Comparison of the proposed KPCA-DeepONet (orange, \textcolor{orange}{{\scriptsize $\blacksquare$}}) and POD-DeepONet (blue, \textcolor{tabblue}{{\Large $\bullet$}}). Lines and shades indicate mean and standard deviation, respectively, over 5 independent trials.}
   \label{fig:errors}
\end{figure}

We compare the proposed KPCA-DeepONet with POD-DeepONet on a 1D nonlinear parametrized function taken from \citep{deim}, the regularized cavity flow \citep{lu2022poddeeponet}, and the Navier--Stokes equation \citep{lu2022poddeeponet}. We evaluate the performance of the networks by computing the $\ell_2$ relative error $\varepsilon$ of the predictions; we perform five independent training trials to compute the mean error and the standard deviation for each test. For POD-DeepONet, we rescale the output as suggested by \cite{lu2022poddeeponet}.~\Cref{fig:errors} summarizes the results for different sizes of the latent space $p$, showing the superior performance of KPCA-DeepONet. For a more detailed comparison and discussion on computational cost, we refer to \cref{app:results,app:cost}, respectively.

\section{Conclusions}

Our results show that employing kernel methods and nonlinear model reduction techniques, i.e. KPCA and kernel ridge regression, combined with DeepONet can provide a more accurate framework for learning operators. The kernel ridge regression employed in KPCA-DeepONet for the reconstruction of the output function can be performed efficiently due to the low dimensionality of the problem in the latent space. Since the reconstruction is nonlinear, our method can be extended for operator learning of PDEs with discontinuities in the future. 

\clearpage


\subsubsection*{Acknowledgements}
All the codes employed for developing KPCA-DeepONet are released as open-source on GitHub-repository~\textcolor{magenta}{\href{https://github.com/HamidrezaEiv/KPCA-DeepONet}{https://github.com/HamidrezaEiv/KPCA-DeepONet}}. Hamidreza Eivazi's research was conducted within the Research Training Group CircularLIB, supported by the Ministry of Science and Culture of Lower Saxony with funds from the program zukunft.niedersachsen of the Volkswagen Foundation.

\subsubsection*{URM Statement}
The authors acknowledge that at least one key author of this work meets the URM criteria of ICLR 2024 Tiny Papers Track.

\bibliography{iclr2023_conference_tinypaper}
\bibliographystyle{iclr2023_conference_tinypaper}
\clearpage
\appendix
\section{Appendix}

\subsection{Deep operator networks (DeepONets)} \label{app:deeponets}
Let us consider a stacked DeepONet with bias \citep{lu2019deeponet}. A DeepONet consists of two sub-networks, i.e., a trunk network and a branch network. The trunk net takes the coordinates as the input and the branch net takes a discretized function $U$ as the input. The operator $\gG$ that maps the input function $U$ to the output function $v$ can be approximated as
\begin{equation}
   \gG(u)(\vy)\approx \sum_{k=1}^{p} b_k(U) t_k(\vy) + b_0,
\end{equation}
for any point $\vy$ in $D^{\prime}$, where $b_0 \in \mathbb{R}$ indicates a bias, $\{b_1, b_2, \cdots, b_p\}$ are the $p$ outputs of the branch net, and $\{t_1, t_2, \cdots, t_p\}$ are the $p$ outputs of the trunk net. The trunk net automatically learns a set of bases for the output function $v$ from the training data. In POD-DeepONet \citep{lu2022poddeeponet}, the trunk net is replaced by a set of POD bases, and the branch net learns their coefficients. Thus, the output can be written as
\begin{equation}
   \gG(u)(\vy)\approx \sum_{k=1}^{p} b_k(U) \phi_k(\vy) + \phi_0(\vy),
\end{equation}
where $\{\phi_1, \phi_2, \cdots, \phi_p\}$ are the POD bases of $v$ and $\phi_0$ is the mean function. 

\subsection{KPCA-DeepONet diagram} \label{app:diagram}

\begin{figure}[h]
   \centerline{\scalebox{0.9}{
  \begin{tikzpicture}[->,>=stealth',shorten >=1pt,auto,node distance=5cm,
                      thick,main node/.style={font=\sffamily\Large\bfseries}]

  \node[main node] (1) {$\gU$};
  \node[main node] (2)  [right of=1] {$\gV$};
  \node[main node] (3) [below of=1,node distance=1.5cm] {$\sR^{n}$};
  \node[main node] (4) [below of=2,node distance=1.5cm,red] {$\sR^{m}$};
  \node[main node] (5) [right of=3,node distance=2.5cm] {$\sR^{p}$};
  
  \path[every node/.style={font=\sffamily\Large\bfseries},blue]
      (1) edge node [above, blue] {$\gG$} (2);

  \path[every node/.style={font=\sffamily\Large\bfseries}]
      (1) edge node [left] {$\gP$} (3);

  \path[every node/.style={font=\sffamily\Large\bfseries}]
      (3) edge node [below] {$b$} (5);
  
  \path[every node/.style={font=\sffamily\Large\bfseries},red]
      (2) edge node [right,red] {$\gQ$} (4);
   
   \path[every node/.style={font=\sffamily\Large\bfseries},red]
   (4) edge node [below,red] {$f$} (5);

   \path[every node/.style={font=\sffamily\Large\bfseries}]
      (5) edge node [above] {$h$} (2);
  
  \end{tikzpicture}}}
  
\caption{A diagram of the KPCA-DeepONet operator learning setup summarizing various maps of interest. $\gG$ is the operator we want to learn. $\gP$ and $\gQ$ are the evaluation operators, $b$ is the mapping by the branch network, $f$ is the projection on the KPCA basis, and $h$ is the mapping by the kernel ridge regression. The red color indicates those mappings that are only required for training.}
\label{fig:diagram}
\end{figure}
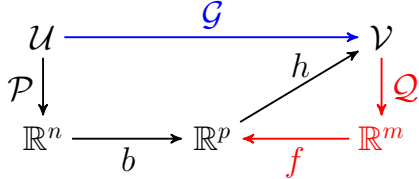

\subsection{The representer theorem.} \label{app:theorem1}
\begin{theorem}
   Representer theorem \citep{learningwithKernels}: Let $\Omega: [0, +\infty) \mapsto \mathbb{R}$ be strictly increasing and let $L$ be a loss function. Consider the optimization problem
   \begin{equation}
      \min_{f \in \gF} L(\vz_i, V_i, f(\vz_i)) + \lambda \Omega(||f||^2_{\gF}),
   \end{equation}
   where $\gF$ is a reproducing kernel Hilbert spaces (RKHS) with kernel $k_z$, and $\lambda > 0$. Then, any optimal solution has the form of $h(\cdot) = \sum_{i = 1}^N \alpha_i k_z(\cdot, \vz_i)$, where $\alpha_i$ are the data-dependent weights.
   \label{theorem1}
\end{theorem}

\subsection{Further experimental results} \label{app:results}

In this section, we present further results obtained from our experiments and compare them to the results reported in \cite{lu2022poddeeponet}. The $\ell_2$ relative errors obtained from the best models are reported in \cref{tab:errors}. Results from \cite{lu2022poddeeponet} correspond to the best model or extension of a model. 

\begin{figure}[h]
   \centering
   \includegraphics[width=0.85\linewidth]{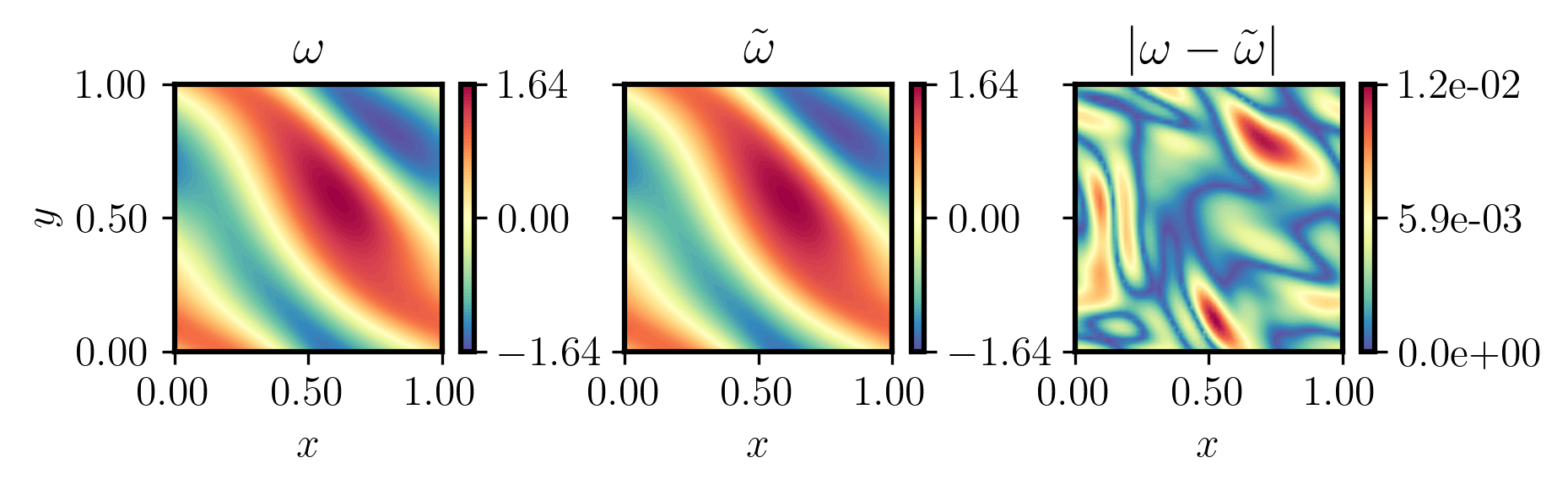}
   \caption{KPCA-DeepONet prediction against the reference data for one sample of the test dataset for the Navier--Stokes equation. $\tilde{\cdot}$ indicates the KPCA-DeepONet prediction.}
   \label{fig:NS}
\end{figure}

\begin{figure}[h]
   \centering
   \includegraphics[width=0.85\linewidth]{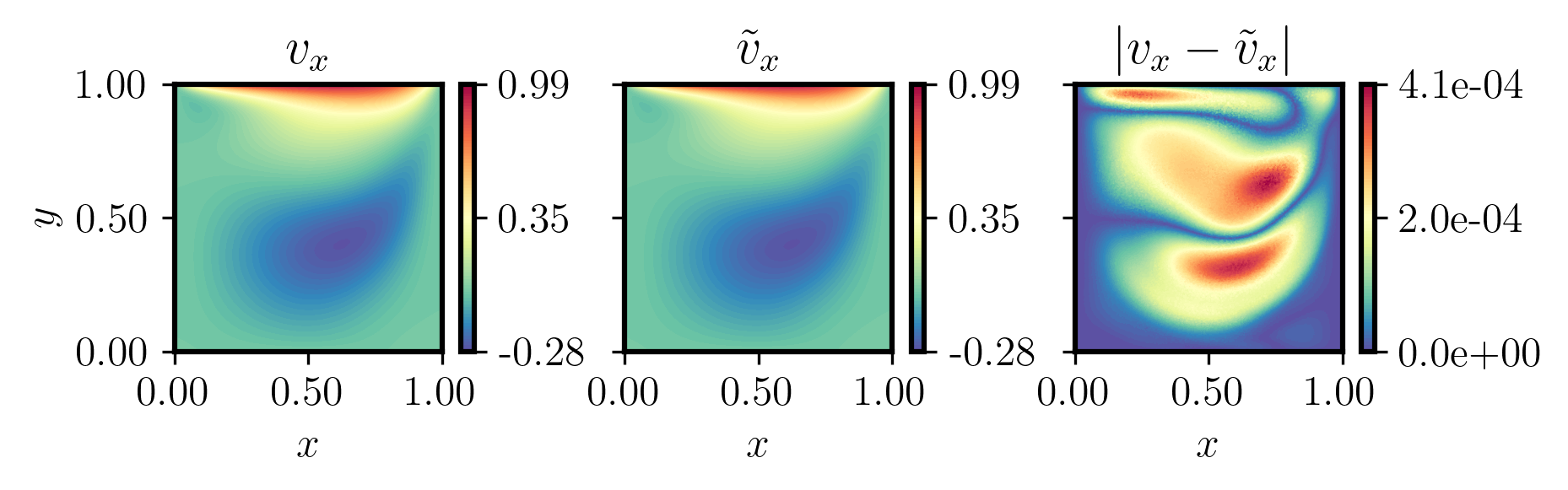}
   \includegraphics[width=0.85\linewidth]{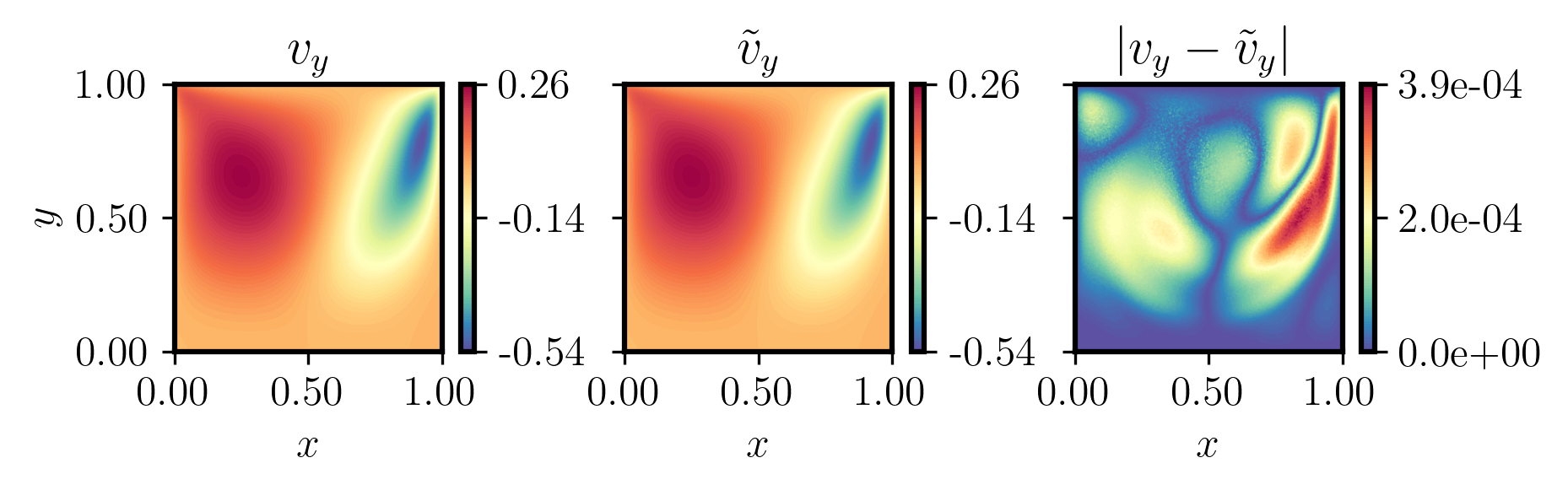}
   \caption{KPCA-DeepONet prediction against the reference data for one sample of the test dataset for the cavity flow. $\tilde{\cdot}$ indicates the KPCA-DeepONet prediction.}
   \label{fig:cavity}
\end{figure}

\Cref{fig:NS,fig:cavity} illustrate the reference data, the prediction of KPCA-DeepONet, and the absolute error of the prediction for the Navier--Stokes and cavity flow problems, respectively. $\tilde{\cdot}$ indicates the KPCA-DeepONet prediction. Note that for the cavity flow problem, the operator network outputs two functions corresponding to the velocity in $x$ and $y$ directions, indicated by $v_x$ and $v_y$, respectively, in \cref{fig:cavity}.
\clearpage

\begin{table}[h]
   \caption{The $\ell_2$ relative errors $\varepsilon$ obtained from different operator networks. Results for models marked with * are taken from \cite{lu2022poddeeponet}.}
   \label{tab:errors}
   \begin{center}
   \begin{tabular}{lccccc}
   Models & 1D nonlinear & Cavity flow & Navier--Stokes \\
   \hline
   KPCA-DeepONet &  \rule{0pt}{3ex} \bm{$0.03 \pm 0.00\%$} &  \bm{$0.05 \pm 0.00\%$} & \bm{$0.96 \pm 0.05\%$} \\
   POD-DeepONet  &   $0.03 \pm 0.00\%$    &  $0.15 \pm 0.03\%$ & $1.15 \pm 0.02\%$ \\
   POD-DeepONet$^*$ &  --     &  $0.33 \pm 0.08\%$ & $1.36 \pm 0.03\%$\\
   DeepONet$^*$ &  --    &  $1.20 \pm  0.23\%$      &  $1.78 \pm 0.02\%$\\
   FNO$^*$ &  --   &  $0.63 \pm 0.04\%$      &  $1.81 \pm 0.02\%$\\
   \hline
   \end{tabular}
   \end{center}
\end{table}

\subsection{Test setup}

The data size of each problem is reported in \cref{tab:data}. We refer readers to section 5.6. of \cite{lu2022poddeeponet} for a detailed description of the problem setup for the Navier--Stokes equation in the vorticity-velocity form, section 5.7. of \cite{lu2022poddeeponet} for the regularized cavity flow (steady) problem, and section 3.3.1. of \cite{deim} for the 1D nonlinear parametrized function. Note that for the 1D nonlinear problem, the goal is to learn a resolution-independent approximation of a parametrized function. The parameters of the function are the inputs of the branch network. 

\begin{table}[h]
   \caption{Dataset size for each problem.}
   \label{tab:data}
   \begin{center}
   \begin{tabular}{lcc}
    & No. of training data & No. of testing data \\
   \hline
   1D nonlinear &  \rule{0pt}{3ex} 51 &  51 \\
   Cavity flow  &   100    &  10 \\
   Navier--Stokes &  1000  &  200 \\
   \hline
   \end{tabular}
   \end{center}
\end{table}

\subsection{Architecture and hyperparameters}

In this section, we provide details on the selected types of kernels and their hyperparameters. We also report the architecture of the branch network.

\begin{table}[h]
   \caption{The selected kernel parameters for each problem.}
   \label{tab:kernels}
   \begin{center}
   \begin{tabular}{lccccccc}   
    & $\gamma_v$ & $c_v$ & $d_v$ & $\gamma_z$ & $c_z$ & $d_z$ & $\lambda$\\
   \hline
   1D nonlinear &  \rule{0pt}{3ex} 1.0 &  0.0 & 1 & 1.0 & 0.0 & 2 & $10^{-3}$\\
   Cavity flow   &  1.0 &  1.0 & 1 & 0.01 & 1.0 & 2 & $10^{-6}$\\
   Navier--Stokes &  1.0 &  0.0 & 1 & $10^{-3}$ & 0.1 & 2 & $10^{-3}$ \\
   \hline
   \end{tabular}
   \end{center}
\end{table}

\begin{figure}[h]
   \centering
   \includegraphics[height=0.2\linewidth]{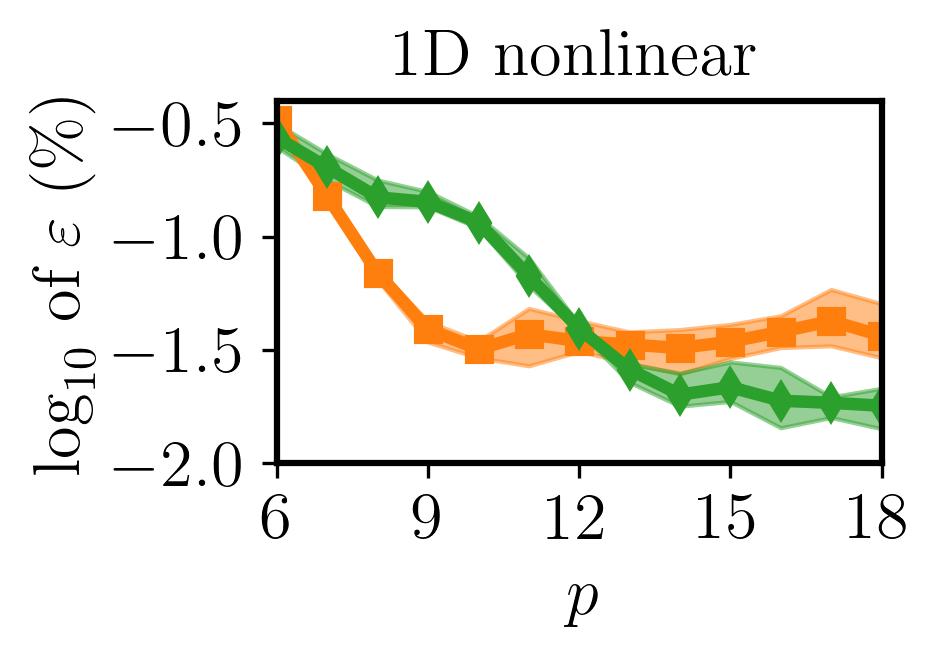}$\qquad$
   \caption{Performance of KPCA-DeepONet when utilizing a linear (orange, \textcolor{orange}{{\scriptsize $\blacksquare$}}) and a quadratic (green, \textcolor{tabgreen}{$\blacklozenge$}) kernel for mapping to the latent space for the 1D nonlinear problem. Lines and shades indicate mean and standard deviation, respectively, over 5 independent trials.}
   \label{fig:dkernel}
\end{figure}

\paragraph{Kernels.} Let $k_v$ and $k_z$ be the kernel functions for KPCA (map from the original space to latent space) and kernel ridge regression (map from the latent space to original space), respectively. For the conducted experiments, we utilize polynomial kernels as
\begin{equation}
   k(\vx, \vy) = (\gamma\, \vx^T \vy + c)^d,
\end{equation}
for simplicity. Alternative kernels, including RBF, Laplace, or Mat\'ern, can also be employed in the proposed context. Note that, in PCA, the reconstruction map involves a direct linear combination of the PCA bases. However, in kernel PCA, the reconstruction map requires a separate step, such as kernel regression, to map the reduced-dimensional representation back to the original input space. The selected kernel parameters and the regularization coefficient for kernel ridge regression $\lambda$ are reported in \cref{tab:kernels} for each problem. Note that we chose a linear kernel for mapping to the latent space in all test cases to underscore the impact of the nonlinear reconstruction on the performance of the learned operator. We compare the performance of KPCA-DeepONet when utilizing both a linear ($d_v = 1$) and a quadratic ($d_v = 2$) kernel for mapping to the latent space for the 1D nonlinear problem. The mapping from the latent space to the output function utilizes a quadratic kernel in both cases. Results are depicted  in \cref{fig:dkernel} for different sizes of the latent space. The results indicate that incorporating nonlinearity could result in better performance when an appropriate size is chosen for the latent space. We obtained $\ell_2$ relative error $\varepsilon$ of $0.02 \pm 0.00\%$ using the quadratic kernel against $0.03 \pm 0.00\%$ using the linear kernel from the best models.

\paragraph{Neural network architecture.} In both KPCA-DeepONet and POD-DeepONet the mapping from the input function to the latent space is performed via the branch network. For both methods, we implement the same architectures and training processes. The branch network architecture for each problem is reported in \cref{tab:branch}. Both the information regarding the output function and its projection into the latent space can be used for training the branch network, with the latter being implemented in this study. The mean-squared error loss for training of the branch network can be expressed as
\begin{equation}
   \gL_b = \dfrac{1}{N_b}\sum_{i=1}^{N_b} \big\Vert \vb_i - \vz_i \big\Vert^2_2, 
\end{equation}
where $\vb_i$ is the output of the branch network for $i$th input sample, $\vz_i$ is the projection of the $i$th output function into the latent space using KPCA, and $N_b$ is the batch size. $||\cdot||_2$ indicates $\ell_2$-norm.

\begin{table}[h]
   \caption{Architecture of the branch network for KPCA-DeepONet and POD-DeepONet for each problem.}
   \label{tab:branch}
   \begin{center}
   \begin{tabular}{lcc}   
    & Branch network & Activation function \\
   \hline
   1D nonlinear &  \rule{0pt}{3ex}Depth 4 \& Width 64 &  tanh \\
   Cavity flow   &  Depth 4 \& Width 64 &  tanh \\
   Navier--Stokes &  CNN &  tanh \\
   \hline
   \end{tabular}
   \end{center}
\end{table}

\paragraph{Training.} For the 1D nonlinear and cavity flow test cases the Adam algorithm is utilized as the optimizer in the training process of the neural network. For the Navier--Stokes problem we employ the AdamW algorithm \citep{adamW}. In all the cases, a scheduled learning rate is used based on the inverse time decay schedule.

\subsection{Computational complexity and cost} \label{app:cost}

In this section, we discuss the computational complexity and cost of the forward maps in KPCA-DeepONet and POD-DeepONet. Both approaches utilize the branch network to map the input function $U$ to the latent vector $\vz$ with the size of $p$. We exclude this step and only discuss the computational complexity of the reconstruction maps from the latent vector to the output function $v$ for one sample. For KPCA-DeepONet the reconstruction map comprises two steps: (1)  computing the kernel matrix $k_z$ with the computational complexity of $\gO(p \times N)$ for a naive implementation, where $N$ is the number of training samples, and (2) mapping to the output function space with the computational complexity of $\gO(N \times m)$, where $m$ is the number of evaluation coordinates. Since $m > p$, we can conclude that the computational complexity of the reconstruction map of the proposed KPCA-DeepONet, for a naive implementation, is $\gO(N \times m)$ while for POD-DeepONet it is $\gO(p \times m)$. The computational complexity of the reconstruction map in KPCA-DeepONet increases linearly with the number of training samples, potentially resulting in memory challenges when dealing with large datasets. Sparse kernel methods may be a suitable remedy for this limitation. 

\begin{figure}[h]
   \centering
   \includegraphics[width=0.55\linewidth]{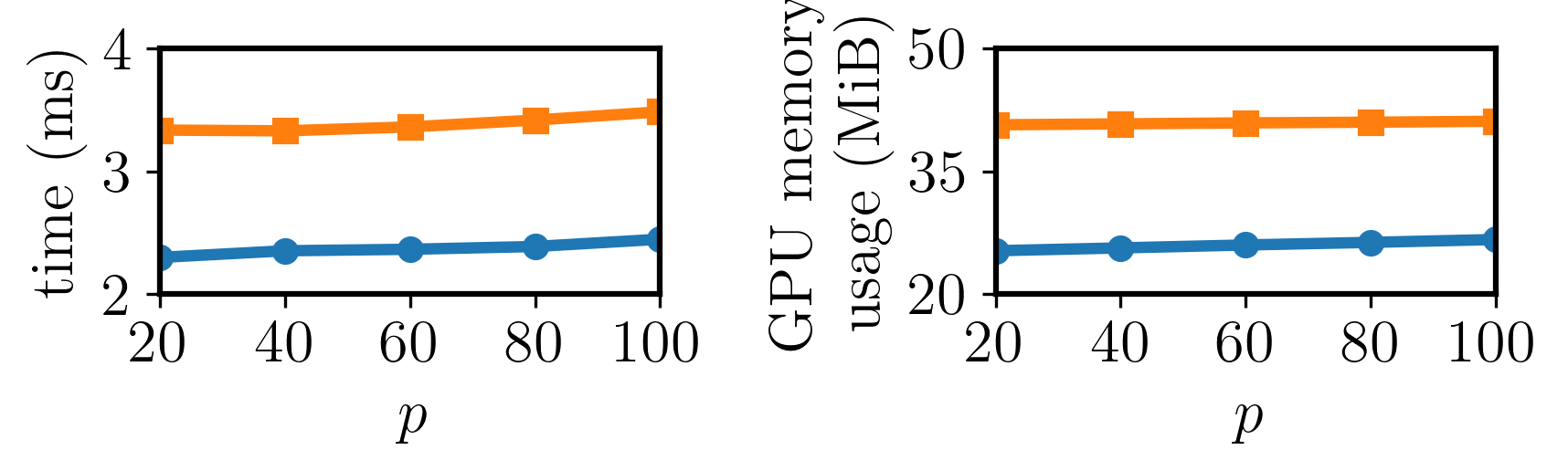}
   \caption{Computational time in milliseconds (ms) (left) and GPU memory usage (right) of the proposed KPCA-DeepONet (orange, \textcolor{orange}{{\scriptsize $\blacksquare$}}) and POD-DeepONet (blue, \textcolor{tabblue}{{\Large $\bullet$}}) versus the size of the latent space. Results are reported for the Navier--Stokes problem.}
   \label{fig:time}
\end{figure}

The computational complexity for modern GPU-based implementations is notably lower than the aforementioned values. We report the computational time and GPU memory usage of the forward map of a batch of 100 samples for the Navier--Stokes problem. Results are depicted for different sizes of the latent space $p$ in \cref{fig:time}. It can be noted that the computational time and GPU memory usage of the KPCA-DeepONet are only marginally higher than those of the POD-DeepONet and do not exhibit scaling with $N/p$. All computations are performed on a workstation with one NVIDIA GeForce RTX 3090 GPU.
 
\end{document}

%% file: math_commands.tex
\usepackage{amsmath,amsfonts,bm}

\def\eqref#1{equation~\ref{#1}}

\def\1{\bm{1}}

\def\vb{{\bm{b}}}

\def\vx{{\bm{x}}}
\def\vy{{\bm{y}}}
\def\vz{{\bm{z}}}

\def\mK{{\bm{K}}}

\DeclareMathAlphabet{\mathsfit}{\encodingdefault}{\sfdefault}{m}{sl}
\SetMathAlphabet{\mathsfit}{bold}{\encodingdefault}{\sfdefault}{bx}{n}

\def\gF{{\mathcal{F}}}
\def\gG{{\mathcal{G}}}

\def\gL{{\mathcal{L}}}

\def\gO{{\mathcal{O}}}
\def\gP{{\mathcal{P}}}
\def\gQ{{\mathcal{Q}}}

\def\gU{{\mathcal{U}}}
\def\gV{{\mathcal{V}}}

\def\sR{{\mathbb{R}}}

